%% file: acl_latex.tex
\pdfoutput=1
\documentclass[11pt]{article}

\usepackage[final]{acl}
\usepackage{acl}
\usepackage{times}
\usepackage{latexsym}
\input{macros.tex}

\title{Structured Tree Alignment for Evaluation of (Speech) Constituency Parsing}
\author{
    Freda Shi \quad Kevin Gimpel \quad Karen Livescu \\
    Toyota Technological Institute at Chicago \\
    6045 S. Kenwood Ave, Chicago, IL, USA, 60637 \\
    \texttt{\{freda,kgimpel,klivescu\}@ttic.edu}
}

\begin{document}
\maketitle
\input{src/00-abstract.tex}
\input{src/10-intro.tex}
\input{src/20-related.tex}

\input{src/30-formulation.tex}
\input{src/40-algorithm.tex}
\input{src/50-example-expr.tex}

\input{src/60-discussion.tex}

\bibliography{anthology,custom}

\newpage
\appendix
\input{src/99-appendix.tex}

\end{document}

%% file: macros.tex
\usepackage[T1]{fontenc}
\usepackage[utf8]{inputenc}
\usepackage{microtype}
\usepackage{amsmath}
\usepackage{amssymb}
\usepackage{amsthm}
\usepackage{bm}
\usepackage{cjhebrew}
\usepackage{xspace}
\usepackage{algpseudocode}
\usepackage{algorithm}
\usepackage{cleveref}
\usepackage{enumitem}
\usepackage{subcaption}
\usepackage{booktabs}
\usepackage{tikz}
\usetikzlibrary{decorations.pathreplacing}
\usepackage[linguistics]{forest}

\newtheorem{theorem}{Theorem}

\newtheorem{corollary}{Corollary}[theorem]
\newtheorem{proposition}[theorem]{Proposition}
\theoremstyle{definition}
\newtheorem{definition}[theorem]{Definition}
\newtheorem{example}{Example}[theorem]
\newtheorem{problem}[theorem]{Problem}
\theoremstyle{remark}

\tikzset{
  mynode/.style={fill,circle,inner sep=1pt,outer sep=0pt}
}

\newcommand{\interalia}[1]{\citep[][\textit{inter alia}]{#1}}

\newcommand{\iou}{\textsc{IoU}\xspace}
\newcommand{\cI}{\mathcal{I}}
\newcommand{\cT}{\mathcal{T}}

\newcommand{\cU}{\mathcal{U}}

\newcommand{\bd}{{\bm{d}}}

\newcommand{\bn}{{\bm{n}}}
\newcommand{\bp}{{\bm{p}}}
\newcommand{\bq}{{\bm{q}}}
\newcommand{\bA}{{\bm{A}}}
\newcommand{\br}{{\bm{r}}}

\newcommand{\bz}{{\bm{z}}}
\newcommand{\structaiou}{\textsc{Struct-IoU}\xspace}
\newcommand{\parseval}{\textsc{ParsEval}\xspace}
\newcommand{\sparseval}{\textsc{SParsEval}\xspace}
\newcommand{\tedeval}{\textsc{TedEval}\xspace}

\DeclareMathOperator*{\mathiou}{\iou}
\DeclareMathOperator*{\suchthat}{\textit{s.t.}}

\algnewcommand\algorithmicforeach{\textbf{for}}
\algdef{S}[FOR]{ForEach}[1]{\algorithmicforeach\ #1\ \algorithmicdo}

\crefname{lemma}{Lemma}{Lemmas}
\crefname{proposition}{Proposition}{Propositions}
\crefname{definition}{Definition}{Definitions}
\crefname{corollary}{Corollary}{Corollaries}
\crefname{example}{Example}{Examples}
\crefname{problem}{Problem}{Problems}
\crefname{remark}{Remark}{Remarks}
\crefname{algorithm}{Algorithm}{Algorithms}
\crefname{figure}{Figure}{Figures}
\crefname{table}{Table}{Tables}
\crefname{section}{\S}{\S\S}
\Crefname{section}{\S}{\S\S}
\crefformat{section}{\S#2#1#3}
\crefname{appendix}{Appendix}{Appendices}
\crefname{equation}{Equation}{Equations}

%% file: src/00-abstract.tex
\begin{abstract}
    We present the structured average intersection-over-union ratio (\structaiou), a similarity metric between constituency parse trees motivated by the problem of evaluating speech parsers.
    \structaiou enables comparison between a constituency parse tree (over automatically recognized spoken word boundaries) with the ground-truth parse (over written words).
    To compute the metric, we project the ground-truth parse tree to the speech domain by forced alignment, align the projected ground-truth constituents with the predicted ones under certain structured constraints, and calculate the average \textsc{IoU} score across all aligned constituent pairs.
    \structaiou takes word boundaries into account and overcomes the challenge that the predicted words and ground truth may not have perfect one-to-one correspondence.
    Extending to the evaluation of text constituency parsing, we demonstrate that \structaiou can address token-mismatch issues, and shows higher tolerance to syntactically plausible parses than \parseval \citep{black-etal-1991-procedure}.\footnote{We open-source the code of \structaiou at \url{https://github.com/ExplorerFreda/struct-iou}.}
\end{abstract}

%% file: src/10-intro.tex
\section{Introduction}
\label{sec:intro}
Automatic constituency parsing of written text \interalia{marcus-etal-1993-building} and speech transcriptions \interalia{godfrey-holliman-1993-switchboard}, as representative tasks of automatic syntactic analysis, have been widely explored in the past few decades.
Appropriate evaluation metrics have facilitated the comparison and benchmarking of different approaches: the \parseval $F_1$ score \citep{black-etal-1991-procedure,sekine1997evalb} has served as a reliable measure of text parsing across scenarios;
for speech transcription parsing, the \sparseval \citep{roark-etal-2006-sparseval} metric extends \parseval and accounts for speech recognition errors by allowing word-level editing with a cost.

\input{src/figs/001-teaser.tex}
Recent work \citep{lai-etal-2023-audio,tseng-etal-2023-cascading} has proposed a new task of textless speech constituency parsing.
In contrast to earlier work that parses manually labeled \interalia{charniak-johnson-2001-edit} or automatic \interalia{kahn-ostendorf-2012-joint} speech transcriptions, these models construct constituency parse trees over automatically recognized spoken word boundaries, where each word is represented with a time range of the spoken utterance, without using any form of text.
To evaluate these textless models, we need a metric that compares the predicted tree (over spoken word boundaries) with the manually labeled ground-truth tree (over written words) and faithfully reflects the parsing quality.
Since the automatically recognized word boundaries may be imperfect, the metric should also reflect the changes in parsing quality due to word boundary errors.
To the best of our knowledge, none of the existing metrics meets these requirements, as they are all designed to compare parse trees over discrete word sequences, instead of continuous time ranges.

Motivated by the need for textless speech parsing evaluation, in this paper, we introduce the structured average intersection-over-union ratio (\structaiou; \cref{fig:example-speech-parse-tree}), a metric that compares two parse trees over time ranges.
We relax the definition of segment trees \citep{bentley1977solutions} to represent speech constituency parse trees, where each node is associated with an interval that represents the time range of the corresponding spoken word or constituent.
To obtain the ``ground-truth'' speech parse trees, we use the forced alignment algorithm \citep{mcauliffe-etal-2017-montreal}, a supervised and highly accurate method that aligns written words to time ranges of the corresponding spoken utterance, to project the ground-truth text parses onto the time domain.
\structaiou is calculated by aligning the same-label nodes in the predicted and ground-truth parse trees, following structured constraints that preserve parent-child relations.
The calculation of \structaiou can be formulated as an optimization problem (\cref{sec:problem-formulation}) with a polynomial-time solution (\cref{sec:solution}) in terms of the number of tree nodes.

Although \structaiou is designed to evaluate speech parsing, it is also applicable to text parsing evaluation.
We analyze \structaiou for both purposes: in speech parsing evaluation, \structaiou robustly takes into account both the structure information and word boundaries; in text parsing evaluation, while maintaining a high correlation with the \parseval $F_1$ score, \structaiou shows a higher tolerance to potential syntactic ambiguity.

%% file: src/figs/001-teaser.tex
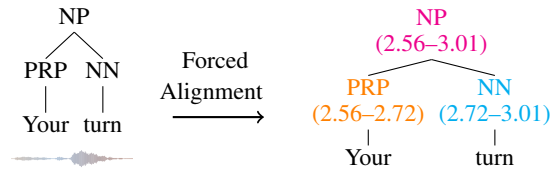
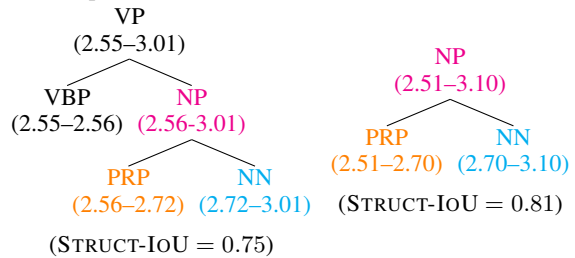
\begin{figure}[t]
    \centering\small
    \begin{subfigure}[t]{0.48\textwidth}
        \centering
        \begin{minipage}{0.23\textwidth}
            \centering
            \begin{forest} for tree={inner sep=0pt,l=10pt,l sep=10pt,s sep=5pt}
                [NP
                    [PRP [Your]]
                    [NN [turn]]
                ]
            \end{forest}
            \includegraphics[width=0.9\textwidth]{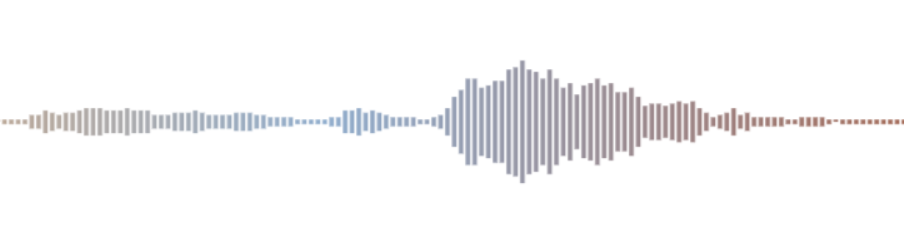}
        \end{minipage}
        \begin{minipage}{0.24\textwidth}
            \centering \normalsize
            {\small Forced \\[-2pt] Alignment}
            \vspace{-8pt}

            \flushleft
            \hspace{0pt}
            \begin{tikzpicture}
                \node[left] at (0,0) (Step 1) {};
                \draw [thick, ->] (Step 1) -- (1.2,0);
            \end{tikzpicture}
        \end{minipage}
        \begin{minipage}{0.48\textwidth}
            \centering
            \begin{forest} for tree={inner sep=0pt,l=10pt,l sep=6pt,s sep=5pt}
                [ \textcolor{magenta}{NP} \\ \textcolor{magenta}{(2.56--3.01)}
                    [ \textcolor{orange}{PRP} \\\textcolor{orange}{(2.56--2.72)}
                        [Your]
                    ]
                    [ \textcolor{cyan}{NN} \\\textcolor{cyan}{(2.72--3.01)}
                        [turn]
                    ]
                ]
            \end{forest}
        \end{minipage}
        \caption{
            \label{fig:example-speech-parse-tree-1}
            Ground-truth speech parse tree (right), obtained by forced alignment between the ground-truth text parse tree (left, top) and the spoken utterance (left, bottom).
        }
    \end{subfigure}
    \begin{subfigure}[t]{0.48\textwidth}
        \centering \small
        \begin{minipage}{0.52\textwidth}
            \centering
            \begin{forest} for tree={inner sep=0pt,l=10pt,l sep=10pt,s sep=5pt}
                [ \textcolor{black}{VP} \\ (2.55--3.01)
                    [ VBP \\ (2.55--2.56) ]
                    [
                        \textcolor{magenta}{NP} \\ \textcolor{magenta}{(2.56-3.01)}
                        [ \textcolor{orange}{PRP} \\ \textcolor{orange}{(2.56--2.72)}]
                        [ \textcolor{cyan}{NN} \\ \textcolor{cyan}{(2.72--3.01)} ]
                    ]
                ]
            \end{forest}

            \vspace{5pt}
            (\structaiou$=0.75$)
        \end{minipage}
        \begin{minipage}{0.45\textwidth}
            \centering
            \begin{forest} for tree={inner sep=0pt,l=10pt,l sep=10pt,s sep=5pt}
                [ \textcolor{magenta}{NP} \\ \textcolor{magenta}{(2.51--3.10)}
                    [ \textcolor{orange}{PRP}\\ \textcolor{orange}{(2.51--2.70)}]
                    [ \textcolor{cyan}{NN} \\ \textcolor{cyan}{(2.70--3.10)} ]
                ]
            \end{forest}

            \vspace{5pt}
            (\structaiou$=0.81$)
        \end{minipage}
        \caption{
            \label{fig:example-speech-parse-tree-2}
            Predicted tree with good word boundaries and an errorful tree structure (left), or that with errorful word boundaries and a perfect tree structure (right).
        }
    \end{subfigure}

    \caption{
            \label{fig:example-speech-parse-tree}
            Illustration of how \structaiou (\cref{sec:problem-formulation,sec:solution}) evaluates textless speech constituency parsing.
            Best viewed in color, where nodes with the same color are aligned.
            Numbers in parentheses are the starting and ending times of the corresponding spans (in seconds).
        }
\end{figure}

%% file: src/20-related.tex
\section{Related Work}

\label{sec:related-work}

\noindent\textbf{Text constituency parsing and evaluation.}
In the past decades, there has been much effort in building and improving constituency parsing models \interalia{collins-koo-2005-discriminative,charniak-johnson-2005-coarse,mcclosky-etal-2006-effective,durrett-klein-2015-neural,cross-huang-2016-span,dyer-etal-2016-recurrent,choe-charniak-2016-parsing,stern-etal-2017-minimal,kitaev-klein-2018-constituency}.
\parseval \citep{black-etal-1991-procedure} has been the standard evaluation metric for constituency parsing in most scenarios, which takes the ground truth and predicted trees and calculates the harmonic mean of precision and recall of labeled spans.
For morphologically rich languages, \tedeval \citep{tsarfaty-etal-2012-joint} extends \parseval to accept multiple morphological analyses over sequences of words.
A few alternative approaches have been pursued to address the potential mismatch in words and sentences \citep{calder-1997-aligning,jo-etal-2024-novel}.
All these metrics are designed to evaluate parses over discrete word sequences, and cannot be easily extended to evaluate speech parses over continuous time ranges.
Although our metric, \structaiou, is designed to evaluate speech constituency parsing, it can be easily extended for text parsing evaluation, reflecting a different aspect from existing metrics (\cref{sec:text-constituency-parsing-evaluation}).

\noindent\textbf{Speech constituency parsing and its evaluation.}
Work on conversational speech parsing has focused on addressing the unique challenges posed by speech, including speech recognition errors \citep{kahn-ostendorf-2012-joint,marin-ostendorf-2014-domain}, unclear sentence boundaries \citep{kahn-etal-2004-parsing}, disfluencies \citep{jamshid-lou-johnson-2020-improving,kahn-etal-2005-effective,lease-johnson-2006-early}, as well as integrating prosodic features into the parsing systems \citep{tran-etal-2018-parsing,tran-ostendorf-2021-assessing}.
On the evaluation side, the closest work to ours is \sparseval \citep{roark-etal-2006-sparseval}, which extends \parseval to account for speech recognition errors by allowing for word-level insertion, deletion, and substitution.
In contrast, our metric \structaiou applies to the cases where no speech recognizer is applied or available.
\vspace{1pt}

\noindent\textbf{Other structured evaluation metrics for parsing.}
There have been evaluation metrics of abstract meaning representations \citep[AMRs;][]{cai-knight-2013-smatch}, where two AMR graphs are matched by solving an NP-complete integer linear programming problem.
While our work shares the spirit with theirs, we focus on the evaluation of speech constituency parsing over continuous word boundaries.
There also exists a polynomial-time exact solution to our optimization problem.

%% file: src/30-formulation.tex
\input{src/figs/01-example.tex}

\section{Preliminaries}
We use real-valued open intervals to represent speech spans for simplicity, although most of the following definitions and conclusions can be easily extended to closed intervals and half-open intervals.
Proof of each corollary and proposition can be found in \cref{appendix:proof}.
\subsection{Open Interval Operations}
\label{sec:open-interval-operations}

\begin{definition}
    \label{def:open-interval-length}
    The \textbf{length} of a real-valued open interval $I=(a, b)$, where $a < b$, is $|I| = b-a$.
\end{definition}

\begin{definition}
    \label{def:intersection-size}
    The \textbf{intersection size} of open intervals $I_1$ and $I_2$ is
    \begin{align*}
        \cI(I_1, I_2) = \left\{
            \begin{aligned}
                & 0 && \text{if } I_1 \cap I_2 = \emptyset \\
                & |I_1\cap I_2| && \text{otherwise.}
            \end{aligned}
            \right.
    \end{align*}
\end{definition}

\begin{definition}
    \label{def:union-size}
    The \textbf{union size} of open intervals $I_1$ and $I_2$ is
    $\cU(I_1, I_2) = |I_1| + |I_2| - \cI(I_1, I_2)$.
\end{definition}

\begin{definition}
    \label{def:intersection-over-union}
    The \textbf{intersection over union} (\iou) ratio between open intervals $I_1$ and $I_2$ is
    \begin{align*}
        \iou(I_1, I_2) = \frac{\cI(I_1, I_2)}{\cU(I_1, I_2)}.
    \end{align*}
\end{definition}
Throughout this paper, we will use \iou as the similarity metric between two intervals.

\subsection{Relaxed Segment Trees}
\label{sec:extended-segment-tree}
We relax the definition of a segment tree \citep{bentley1977solutions} as follows to represent parse trees.

\begin{definition}
    \label{def:segment-tree-node}
    A \textbf{node} $\bn$ of a relaxed segment tree is a triple
    $\bn = \langle
        I_{\bn},
        C_{\bn},
        \ell_{\bn}
    \rangle$, where $\ell_{\bn}$ refers to the label of the node, and 
    \begin{enumerate}[leftmargin=*,topsep=2pt, noitemsep]
        \item $I_{\bn} = (s_{\bn}, e_{\bn})$ is an open interval (i.e., segment) associated with the node $\bn$, where $s_{\bn} < e_{\bn}$;
        \item $C_{\bn}$ is a finite set of disjoint children nodes of $\bn$: for any $\bp, \bq \in C_\bn (\bp \neq \bq)$, $I_\bp \cap I_\bq = \emptyset$.
        $C_{\bn} = \emptyset$ if and only if $\bn$ is a terminal node;
        \item For a nonterminal node $\bn$, $s_{\bn} = \min_{\bp\in C_{\bn}} s_{\bp},$ and $e_{\bn} = \max_{\bp\in C_{\bn}} e_{\bp}$.
    \end{enumerate}
\end{definition}
\begin{corollary}
    \label{corollary:segment-tree-parent-cover}
    For nodes $\bp, \bn$, if $\bp \in C_\bn$, then $I_\bp \subseteq I_\bn$.
\end{corollary}

\begin{definition}
    \label{def:ancestor}
    Node $\bp$ is an \textbf{ancestor} of node $\bq$ if there exists a sequence of nodes $\bn_0,$ $\bn_1,$ $\ldots,$ $\bn_k (k\geq 1)$ such that (i.) $\bn_0 = \bp$, (ii.) $\bn_k = \bq$, and (iii.) for any $i \in [k]$,\footnote{$[k]=\{1, 2, \ldots, k\}$, where $k \in \mathbb{N}$.} $\bn_{i} \in C_{\bn_{i-1}}$.
\end{definition}

\begin{corollary}
    \label{corollary:ancestor-cover}
    If node $\bp$ is an ancestor of node $\bq$, then $I_\bp \supseteq I_\bq$.
\end{corollary}

\begin{definition}
    \label{def:descendant}
    Node $\bp$ is a \textbf{descendant} of node $\bq$ if $\bq$ is an ancestor of $\bp$.
\end{definition}

\begin{definition}
    \label{def:ex-segment-tree}
    A \textbf{relaxed segment tree} $\cT = \langle\br_\cT, N_\cT\rangle$ is a tuple, where
    \begin{enumerate}[leftmargin=*,topsep=2pt,noitemsep]
        \item $\br_\cT$ is the root node of $\cT$;
        \item $N_\cT = \{\br_\cT\} \cup \{\bn: \bn \text{ is a descendant of } \br_\cT\}$ is a finite set of all nodes in $\cT$.
    \end{enumerate}
\end{definition}
\begin{example}
    A constituency parse tree over spoken word time ranges (\cref{fig:example-speech-parse-tree-1}) can be represented by a relaxed segment tree.
\end{example}

\begin{corollary}
    \label{corollary:segment-tree-characterization}
    A relaxed segment tree can be uniquely characterized by its root node.
\end{corollary}
\noindent In the following content, we use $\cT(\bn)$ to denote the relaxed segment tree rooted at $\bn$.

\begin{proposition}
    \label{proposition:non-ancestor-disjoint}
    For a relaxed segment tree $\cT$ and $\bp, \bq \in N_\cT$,  $\bp$ is neither an ancestor nor a descendant of $\bq$ $\Leftrightarrow$ $I_\bp \cap I_\bq = \emptyset$.
\end{proposition}

\section{The \structaiou Metric}
\label{sec:structaiou}
\subsection{Problem Formulation}
\label{sec:problem-formulation}
Given relaxed segment trees $\cT_1$ and $\cT_2$ with node sets $N_{\cT_1} = \{\bn_{1, i}\}_{i=1}^{|N_{\cT_1}|}$ and $N_{\cT_2} = \{\bn_{2, j}\}_{j=1}^{|N_{\cT_2}|}$, we can align the trees by matching their same-label nodes.
Let $\bn_{1,i}\leftrightarrow \bn_{2,j}$ denote the matching between the nodes $\bn_{1,i}$ and $\bn_{2,j}$.

\begin{definition}[conflicted node matchings; \cref{fig:example-conflicted-node-matching}]
    \label{def:conflicted-node-matching}
    The matchings $\bn_{1, i} \leftrightarrow \bn_{2, j}$ and $\bn_{1, k} \leftrightarrow \bn_{2, \ell}$ are \textit{conflicted} if any of the following conditions holds:
    \begin{enumerate}[leftmargin=*, topsep=2pt, noitemsep]
        \item $\bn_{1,i}$ is an ancestor of $\bn_{1,k}$, and $\bn_{2,j}$ is not an ancestor of $\bn_{2,\ell}$;
        \item $\bn_{1,i}$ is not an ancestor of $\bn_{1,k}$, and $\bn_{2,j}$ is an ancestor of $\bn_{2,\ell}$;
        \item $\bn_{1,i}$ is a descendant of $\bn_{1,k}$, and $\bn_{2,j}$ is not a descendant of $\bn_{2,\ell}$;
        \item $\bn_{1,i}$ is not a descendant of $\bn_{1,k}$, and $\bn_{2,j}$ is a descendant of $\bn_{2,\ell}$.
    \end{enumerate}
\end{definition}
\noindent Intuitively, we would like the alignment to be consistent with the ancestor-descendant relationship between nodes.

The optimal (i.e., maximally \iou-weighted) structured alignment between $\cT_1$ and $\cT_2$ is given by the solution to the following problem:
\begin{problem}[maximally \iou-weighted alignment]
\label{prob:max-iou-weighted-tree-alignment}
\begin{align*}
    & \bA^* = \arg\max_{\bA}
        \quad \sum_{i=1}^{|N_{\cT_1}|}\sum_{j=1}^{|N_{\cT_2}|} a_{i,j}
        \mathiou\left(
            I_{\bn_{1, i}},
            I_{\bn_{2, j}}
        \right)
\end{align*}
\begin{align}
    \suchthat
        & \sum_{j} a_{i,j} \leq 1 (\forall i \in \left[\left|N_{\cT_1}\right|\right]) \label{eq:max-iou-weighted-tree-alignment-constraint1}, \\
        & \sum_{i} a_{i,j} \leq 1 (\forall j \in \left[\left|N_{\cT_2}\right|\right]) \label{eq:max-iou-weighted-tree-alignment-constraint2}, \\
        & a_{i,j} + a_{k,\ell} \leq 1 \nonumber \text{if } \bn_{1, i}\leftrightarrow \bn_{2,j} \text{ and }  \bn_{1, k}\leftrightarrow \bn_{2,\ell} \\& \text{ are conflicted.} \nonumber
\end{align}
\end{problem}
\noindent $\bA\in \{0, 1\}^{|N_{\cT_1}|\times|N_{\cT_2}|}$ denotes an alignment matrix: $a_{i, j} = 1$ indicates that the matching $\bn_{1, i} \leftrightarrow \bn_{2, j}$ is selected, otherwise $a_{i, j} = 0$.
The last constraint of \cref{prob:max-iou-weighted-tree-alignment} ensures that there are no conflicted matchings selected.
\cref{eq:max-iou-weighted-tree-alignment-constraint1,eq:max-iou-weighted-tree-alignment-constraint2} imply one-to-one matching between nodes; that is, in a valid tree alignment, each node in $\cT_1$ can be matched with at most one node in $\cT_2$, and vice versa.
The solution to \cref{prob:max-iou-weighted-tree-alignment} gives the maximal possible sum of \iou over aligned node pairs.

\begin{definition}
    \label{def:structa-iou}
    The \textbf{structured average} \textbf{\iou} \\
    (\structaiou) between $\cT_1$ and $\cT_2$ is given by
    \begin{align*}
        & \overline{\iou}(\cT_1, \cT_2) \\
        = & \frac{1}{|N_{\cT_1}| + |N_{\cT_2}|} \sum_{i=1}^{|N_{\cT_1}|}\sum_{j=1}^{|N_{\cT_2}|} a^*_{i,j} \mathiou\left(I_{\bn_{1, i}}, I_{\bn_{2, j}}\right),
    \end{align*}
    where $\bA^* = \{a^*_{i,j}\}$ is the solution to Problem~\ref{prob:max-iou-weighted-tree-alignment}.
\end{definition}

%% file: src/figs/01-example.tex
\begin{figure*}[t]
    \centering
    \begin{minipage}[t]{0.15\textwidth}
        $\cT_1$:
        \begin{forest} for tree={inner sep=0pt,l=10pt,l sep=10pt,s sep=15pt}
            [A
                [B]
                [C
                    [D]
                    [E]
                ]
            ]
        \end{forest}
    \end{minipage}
    \begin{minipage}[t]{0.15\textwidth}
        $\cT_2$:
        \begin{forest} for tree={inner sep=0pt,l=10pt,l sep=10pt,s sep=15pt}
            [F
                [G]
                [H]
            ]
        \end{forest}
    \end{minipage}\hspace{0.05\textwidth}
    \begin{minipage}[t]{0.6\textwidth}
        \vspace{-24pt}
        \begin{align*}
            \text{A} \leftrightarrow \text{G} \text{ and } &\text{E} \leftrightarrow \text {H:}&&\text{conflicted (violating rule 10.1)} \\
            \text{C} \leftrightarrow \text{F} \text{ and } &\text{A} \leftrightarrow \text {H:}&&\text{conflicted (violating rules 10.2 and 10.3)} \\
            \text{B} \leftrightarrow \text{H} \text{ and } &\text{C} \leftrightarrow \text {F:}&&\text{conflicted (violating rule 10.4)} \\
            \text{A} \leftrightarrow \text{F} \text{ and } &\text{C} \leftrightarrow \text {H:}&&\text{not conflicted}
        \end{align*}
    \end{minipage}
    \caption{
        \label{fig:example-conflicted-node-matching}
        Examples of conflicted and non-conflicted node matchings (\cref{def:conflicted-node-matching}).
    }
\end{figure*}
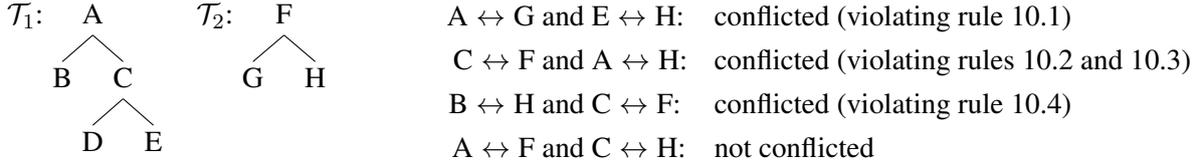

%% file: src/40-algorithm.tex
\subsection{Solution}
\label{sec:solution}

\begin{algorithm*}[t]
    \caption{Polynomial time solution to \cref{eq:subspan-alignments} \label{algo:p-solution}}
    \begin{algorithmic}
    \Require $\cT_1, \cT_2$\;
    \State $g[\cT_1, \cT_2, x, y] \gets 0, \forall x, y$\;
    \State $g'[\cT_1, \cT_2, x, y] := \max_{x' < x, y' < y} g[\cT_1, \cT_2, x', y'] $\;
    \State $\bd_1 \gets$ the sequence of all descendants of $\br_{\cT_1}$, sorted in increasing order of right endpoint\;
    \State $\bd_2 \gets$ the sequence of all descendants of $\br_{\cT_2}$, sorted in increasing order of right endpoint\;
    \ForEach{$i \gets 1 \ldots, |\bd_1|$}
        \ForEach{$j \gets 1 \ldots, |\bd_2|$}
            \State $g[\cT_1, \cT_2, e_{\bd_{1, i}}, e_{\bd_{2, j}}] \gets \max(g[\cT_1, \cT_2, e_{\bd_{1, i}}, e_{\bd_{2, j}}], f_{\cT(\bd_{1, i}), \cT(\bd_{2, j})} + g'[\cT_1, \cT_2, s_{\bd_{1, i}}, s_{\bd_{2, j}}])$
            \State update $g'$ accordingly within $\mathcal{O}(1)$ time using bidimensional prefix sum 
        \EndFor
    \EndFor
    \Ensure \cref{eq:subspan-alignments} $ = g[\cT_1, \cT_2, e_{\br_{\cT_1}}, e_{\br_{\cT_2}}]$
    \end{algorithmic}
\end{algorithm*}

We present a polynomial-time algorithm for the exact solution to \cref{prob:max-iou-weighted-tree-alignment}, by breaking it down into structured subproblems and solving them recursively with dynamic programming.

We define the subproblem as follows: given relaxed segment trees $\cT_1$ and $\cT_2$, we would like to find the maximum \iou weighted alignment of $\cT_1$ and $\cT_2$, where the roots of $\cT_1$ and $\cT_2$ are aligned.
Without loss of generality, we assume that the root nodes of $\cT_1$ and $\cT_2$ are both indexed by $1$.
Formally,
\begin{problem}[maximum \iou weighted alignment, with root nodes aligned]
    \label{prob:root-aligned-max-iou-weighted-tree-alignment}
    \begin{align*}
        f_{\cT_1, \cT_2} = \max_{\bm{A}} \sum_{i=1}^{|N_{\cT_1}|}\sum_{j=1}^{|N_{\cT_2}|} a_{i,j}
        \mathiou\left(
            I_{\bn_{1, i}},
            I_{\bn_{2, j}}
        \right)
    \end{align*}

    \vspace{-20pt}

    \begin{align*}
    \suchthat ~
        & a_{1, 1} = 1; \\
        & \sum_{j} a_{i,j} \leq 1 (\forall i \in \left[\left|N_{\cT_1}\right|\right]), \\
        & \sum_{i} a_{i,j} \leq 1 (\forall j \in \left[\left|N_{\cT_2}\right|\right]), \\
        & a_{i,j} + a_{k,\ell} \leq 1 \text{if } \bn_{1, i}\leftrightarrow \bn_{2,j} \text{ and} \bn_{1,k}\leftrightarrow \bn_{2,\ell} \\ &\text{ are conflicted},
    \end{align*}
    where $\bA\in \{0, 1\}^{|\cT_1|\times|\cT_2|}$ is the alignment matrix.
\end{problem}
While \cref{prob:max-iou-weighted-tree-alignment,prob:root-aligned-max-iou-weighted-tree-alignment} are not equivalent in principle, \cref{prob:max-iou-weighted-tree-alignment} can be reduced to \cref{prob:root-aligned-max-iou-weighted-tree-alignment} within $\mathcal{O}(1)$ time, by adding a dummy root node to each tree that associates with segments covering all the segments in both trees.
We now present a polynomial-time solution to \cref{prob:root-aligned-max-iou-weighted-tree-alignment}.

\begin{definition}
    \label{def:ordered-disjoint-descendant-sequence}
    Given a node $\bn$ of a relaxed segment tree, $\mathbf{D} = \left(\bn_1, \bn_2, \ldots, \bn_k\right)$ is an \textbf{ordered disjoint descendant sequence} of $\bn$ if
    \begin{enumerate}[leftmargin=*, topsep=2pt]
        \setlength{\itemsep}{0pt}
        \item (ordered) for any $i, j \in [k]$ and $i < j$, $s_{\bn_i} < s_{\bn_j}$, where $s_{\bn_i}$ and $s_{\bn_j}$ are left endpoint of the associated intervals;
        \item (disjoint) for any $i, j \in [k]$ and $i\neq j$,
        $I_{n_i} \cap I_{n_j} = \emptyset$;
        \item (descendant) for any $i \in [k]$, $\bn_i$ is a descendant of $\bn$.
    \end{enumerate}
\end{definition}
\begin{corollary}
    \label{cor:ordered-disjoint-descendant-sequence}
    In an ordered disjoint descendant sequence $\mathbf{D} = \left(\bn_1, \bn_2, \ldots, \bn_k\right)$ of $\bn$, $e_{\bn_i} \leq s_{\bn_{i+1}}$ for any $i \in [k-1]$.
\end{corollary}

The solution to \cref{prob:root-aligned-max-iou-weighted-tree-alignment} is given by the following recursion:
\begin{align}
    f_{\cT_1, \cT_2} = & \iou\left(I_{\br_{\cT_1}}, I_{\br_{\cT_2}}\right) + \nonumber \\
    &
    \max_{|\mathbf{D}_1|=|\mathbf{D}_2|}
        \sum_{i=1}^{|\mathbf{D}_1|} f_{\cT(\bd_{1, i}), \cT(\bd_{2, i})} \label{eq:subspan-alignments},
\end{align}
where $\br_{\cT_1}$ and $\br_{\cT_2}$ denote the root nodes of $\cT_1$ and $\cT_2$ respectively; $|\cdot|$ denotes the length of a sequence; $\mathbf{D}_1 = (\bd_{1, 1}, \bd_{1, 2}, \ldots, \bd_{1, |\mathbf{D}_1|})$ and $\mathbf{D}_2 = (\bd_{2, 1}, \bd_{2, 2}, \ldots, \bd_{2, |\mathbf{D}_2|})$ are same-length ordered disjoint descendant sequences of $\br_{\cT_1}$ and $\br_{\cT_2}$ respectively.
\cref{eq:subspan-alignments} can be computed within polynomial time, by solving a knapsack-style problem with dynamic programming. Specifically, let
\begin{align*}
    g[\cT_1, \cT_2, &e_1, e_2] = \\
    & \max_{|\mathbf{D}_1^{e_1}|=|\mathbf{D}_2^{e_2}|}
    \sum_{j=1}^{|\mathbf{D}_1^{e_1}|} f_{\cT(\bd_{1, j}^{e_1}), \cT(\bd_{2, j}^{e_2})},
\end{align*}
where $e_1$ and $e_2$ are arbitrary scalars denoting the constraints of endpoints;
$\mathbf{D}_1^{e_1} = (\bd_{1, 1}^{e_1}, \ldots, \bd_{1, |\mathbf{D}_1^{e_1}|}^{e_1})$ is an ordered disjoint descendant sequence of $\br_{\cT_1}$, where for any $j \in [|\mathbf{D}_1^{e_1}|]$, the right endpoint of the corresponding node $e_{\bd_{1, j}^{e_1}} \leq e_1$;
similarly, $\mathbf{D}_2^{e_2} = (\bd_{2, 1}^{e_2}, \bd_{2, 2}^{e_2}, \ldots, \bd_{2, |\mathbf{D}_2^{e_2}|}^{e_2})$ is a disjoint descendant sequence of $\br_{\cT_2}$ of which the right endpoint of each node does not exceed $e_2$.
\cref{algo:p-solution} computes $g$ and \cref{eq:subspan-alignments} within polynomial time, and therefore leads to a polynomial-time solution to \cref{prob:root-aligned-max-iou-weighted-tree-alignment}.

\paragraph{Complexity analysis.}
Suppose $|\cT_1| = n$ and $|\cT_2| = m$.
To compute $f_{\cT_1, \cT_2}$, all we need to compute is $g[\cT_1', \cT_2', e_1', e_2']$ for all $\cT_1', \cT_2'$, $e_1'$ and $e_2'$.
Here, $\cT_1'$ and $\cT_2'$ enumerate over all subtrees of $\cT_1$ and $\cT_2$, respectively, and $e_1'$ and $e_2'$ enumerate over the endpoints of all nodes in both trees, respectively.
The update process requires $\mathcal{O}(1)$ time for each $\cT_1, \cT_2, e_1, e_2$.
The edge cases, i.e., $g$ values of terminal nodes, can be directly computed in $\mathcal{O}(1)$ time, and therefore, the overall time complexity to solve \cref{prob:root-aligned-max-iou-weighted-tree-alignment} is $\mathcal{O}(n^2m^2)$.

%% file: src/50-example-expr.tex
\section{Experiments}
\label{sec:experiments}
\subsection{Speech Constituency Parsing Evaluation}
\label{sec:speech-constituency-parsing-evalution}

We use the NXT-Switchboard \citep[NXT-SWBD;][]{calhoun-etal-2010-nxt} dataset to train and evaluate models, where the parser can access the forced alignment word boundaries in both training and testing stages.
We train an off-the-shelf supervised constituency parsing model for speech transcriptions \citep{jamshid-lou-johnson-2020-improving} on the training set of NXT-SWBD, do early-stopping using \parseval $F_1$ on the development set, and perform all the analysis below on the development set.
The model achieves $F_1=85.4$ and \structaiou (averaged across sentences)\footnote{Unless otherwise specified, all \structaiou scores reported in the paper are computed by averaging across \structaiou scores of individual sentences. We compare and discuss sentence-level and corpus-level \structaiou in \cref{sec:sentence-vs-corpus-structaiou}}$=0.954$ on the standard development set.

\input{src/figs/02-speech-comparison.tex}
\subsubsection{Comparison with \parseval $F_1$}
\label{subsec:comparison-to-parseval-f1-speech}
Since the forced alignment word boundaries are accessible by the models, the \parseval $F_1$ metric can be directly calculated between the predicted speech constituency parse tree and the ground truth.
We compare the values of \structaiou and \parseval (\citet{sekine1997evalb} implementation with default parameters) in the settings with forced-alignment word segmentation (\cref{fig:structa-iou-vs-parseval-speech}), and find a strong correlation between the two metrics.

\subsubsection{Analysis: \structaiou with Perturbed Word Boundaries}
\label{subsec:perturbation-speech}
In textless speech parsing \citep{lai-etal-2023-audio,tseng-etal-2023-cascading}, the word boundaries are unknown, and the boundaries predicted by the parser are usually imperfect.
As a controlled simulation to such settings, we perturb the forced alignment word boundaries of the predicted parse tree (\cref{fig:perturbation-example}), and calculate the \structaiou score between the perturbed parse tree and the ground truth over the original forced alignment word boundaries.
Specifically, we suppose the word boundaries of a sentence with $n$ words are $\mathcal{B}=b_0, b_1, \dots, b_n$,\footnote{We assume no silence between spoken words; if any inter-word silence exists, we remove it.} and consider the following types of perturbation with a hyperparameter $\delta \in [0, 1]$ controlling the perturbation level:
\begin{itemize}[leftmargin=*,topsep=2pt,noitemsep]
    \item \textbf{Noise}-$\delta$. We start with $\mathcal{B}^{(0)} = \mathcal{B}$, and update the boundaries iteratively as follows.
    For each $i \in [n-1]$, we randomly draw a number $r_i$ from the uniform distribution $U(-\delta, \delta)$, and let $b_i^{(i)} = b_i^{(i-1)} + |r_i| * \left(b_{i+\text{sgn}(r_i)}^{(i-1)} - b_i^{(i-1)}\right)$, where $\text{sgn}(\cdot): \mathbb{R}\rightarrow \{1, -1\}$ denotes the sign function
    \begin{align*}
        \text{sgn}(x) = \begin{cases}
            1 & \text{if } x \geq 0; \\
            -1 & \text{if } x < 0.
        \end{cases}
    \end{align*}
    For all $j\neq i$ and $j\in [n]$, we let $b_j^{(i)}$ remain the same as $b_j^{(i-1)}$.
    Finally, we take $\mathcal{B}^{(n-1)}$ as the perturbed word boundaries for the predicted tree.
    \item \textbf{Insert}-$\delta$. We randomly draw a number $r_i$ from the uniform distribution for each boundary index $i\in[n]$.
    If $r_i < \delta$, we insert a word boundary at the position $b'_i$, randomly drawn from the uniform distribution $U(b_{i-1}, b_i)$, breaking the $i^{\textit{th}}$ spoken word into two (i.e., $[b_{i-1}, b'_i]$ and $[b'_i, b_i]$).
    \item \textbf{Delete}-$\delta$. Similarly to the insertion-based perturbation, we randomly draw a number $r_i$ from the uniform distribution $U(0, 1)$ for each boundary index $i \in [n-1]$, and delete the boundary $b_i$ if $r_i < \delta$.
    Since such boundary deletion may break the predicted tree structure, we use the base model to re-predict the parse tree with the new word boundaries, where words concatenated by space are taken as the textual input \citep{jamshid-lou-johnson-2020-improving}.
\end{itemize}
A larger $\delta$ means a higher level of perturbation is applied, and we therefore expect a lower \structaiou score; $\delta=0$ means no perturbation is applied, and the \structaiou score is the same as that for the predicted parse trees with forced alignment word boundaries.
\input{src/figs/03-speech-perturbation-example.tex}

For each $\delta \in \{0.1, 0.2, \ldots, 1.0\}$, starting from the base model (for deletion-based perturbation) or its predicted parse trees (for noise and insertion-based perturbation), we run the perturbation 5 times and report both the mean and the standard deviation of the \structaiou result after perturbation.

\paragraph{Results and discussion.}
We present how the \structaiou value changes with respect to $\delta$ for different types of perturbation (\cref{fig:perturbation-speech}).
The standard deviation is nearly invisible in the figure, showing that our metric is stable under a specific setting.
For all three types of perturbation, as desired, a larger $\delta$ leads to a lower \structaiou score.
Among the perturbation types, \structaiou is the most sensitive to deletion, and the least sensitive to noise-based perturbation.
Although the results are not comparable across perturbation types in the most rigorous sense, this reflects the fact that \structaiou, to some extent, is more sensitive to structural change of the trees than simple word boundary changes.

Although both word boundary insertion and deletion change the predicted tree structures, the former has less impact on the \structaiou scores.
This also aligns with our expectation: boundary insertion only splits some of the spoken words into two and keeps the longer constituents; however, deletion may change significantly the tree structure, especially when it happens at the boundary of two long constituents.
\input{src/figs/04-speech-perturbation-results.tex}

\subsubsection{Corpus-Level vs. Sentence-Level Metric}
\label{sec:sentence-vs-corpus-structaiou}
Note that 39.7\% utterances in the NXT-SWBD development set contain only one spoken word, and the \structaiou score of such sentences is always high---the metric degenerates to the \iou score between two intervals.
Averaging the \structaiou scores across all sentence pairs in the dataset may therefore overly emphasize these short utterances.
To address this, we introduce the corpus-level \structaiou score as an alternative, where \cref{def:structa-iou} is modified as follows:
\begin{definition}
    \label{def:structa-iou-corpus}
    The corpus-level \structaiou between two sets of parsed trees $\mathcal{D}_1 = \{\cT_{1,k}\}$ and $\mathcal{D}_2 = \{\cT_{2,k}\}$ is given by
    \begin{align*}
        & \overline{\iou}(\mathcal{D}_1, \mathcal{D}_2) \\
        = & \frac{\sum_{k=1}^{|\mathcal{D}_1|} \left(|\cT_{1,k}| + |\cT_{2,k}|\right)\overline{\iou}(\cT_{1,k}, \cT_{2,k})}{\sum_{k'=1}^{|\mathcal{D}_1|} |\cT_{1,k'}| + |\cT_{2,k'}|},
    \end{align*}
    where $|\mathcal{D}_1| = |\mathcal{D}_2|$, and a pair of $\cT_{1,k}$ and $\cT_{2,k}$ denotes the parse trees of the $k^{\textit{th}}$ sentence in the corpus respectively.
\end{definition}
\input{src/figs/05-speech-structaiou-with-length.tex}

We compare the corpus-level and sentence-level \structaiou scores (\cref{fig:structa-iou-with-length}).
As desired, the corpus-level \structaiou score has lower absolute values than the sentence-level one, and the difference is more significant when longer sentences are considered.
A similar phenomenon has been found in text constituency parsing \citep{kim-etal-2019-compound} as well, where corpus-level \parseval $F_1$ scores are lower than sentence-level ones.

\subsection{Evaluation of English Text Constituency Parsing}
\label{sec:text-constituency-parsing-evaluation}
We extend our experiment to the evaluation of text constituency parsing.
In this part, we suppose every written word corresponds to a unit-length segment---analogously, this can be considered as speech parsing with evenly distributed word boundaries, for both predicted and ground-truth trees.

\subsubsection{Correlation with \parseval $F_1$ Scores on the Penn Treebank}
We use the Penn Treebank \citep[PTB;][]{marcus-etal-1993-building} dataset to train and evaluate Benepar \citep{kitaev-klein-2018-constituency}, a state-of-the-art text constituency parsing model, doing early-stopping using labeled \parseval $F_1$ on the development set.
The base model achieves \parseval $F_1=94.4$ and \structaiou (averaged across sentences) $=0.962$ on the standard development set.

\input{src/figs/06-text-comparison.tex}
We compare the \structaiou scores with the \parseval $F_1$ scores on the development set (\cref{fig:structa-iou-vs-parseval-text}).
As in the speech parsing experiment, we find a strong correlation between the two metrics, showing that \structaiou is consistent with the existing metric in the text parsing domain.

\subsubsection{\structaiou vs. \parseval $F_1$ on Syntactically Ambiguous Sentences}
\label{sec:syntactically-ambiguous-sentences}
\input{src/figs/07-syntactic-ambiguity.tex}
We consider a special setting of parsing syntactically ambiguous sentences, where the syntactically plausible parse tree of a sentence may not be unique (see examples in \cref{fig:example-ambiguous-sentences}).
We simplify the case shown in \cref{fig:example-ambiguous-sentences} using synthetic sentences with syntactic ambiguity with the template \texttt{N (P N)\{{\it n}\}}, where \texttt{P} denotes a preposition and \texttt{N} denotes a noun, and \texttt{{\it n}} determines how many times the \texttt{P N} pattern is repeated.
For \texttt{N (P N)\{2\}}, the two potential parse trees are shown in \cref{fig:example-ambiguous-sentences-synthetic}.

We compare \parseval and \structaiou in the following scenarios, choosing a random syntactically plausible parse tree as the ground truth:
\begin{itemize}[leftmargin=*,topsep=2pt,noitemsep]
    \item \textbf{Ground truth vs. random parse trees}, where the random parse trees are constructed by recursively combining random consecutive words (or word groups) into a binary tree. We construct 100 random parse trees and report the average.
    \item \textbf{Ground truth vs. syntactically plausible parse trees}, where we report the lowest possible score between the ground truth and other syntactically plausible trees.
\end{itemize}
As shown in \cref{tab:syntactic-ambiguity-results}, the lowest possible \parseval $F_1$ score between the ground truth and another syntactically plausible tree is significantly lower than the score achieved by meaningless random trees; however, \structaiou consistently assigns higher scores to the syntactically plausible parses, showing a higher tolerance to syntactic ambiguity.

\input{src/figs/08-syntactic-ambiguity-results.tex}

\subsection{Evaluation of Hebrew Text Constituency Parsing}
\label{sec:hebrew-parsing}
We extend our experiment to the evaluation of text constituency parsing on Hebrew, a morphologically rich language that allows different tokenizations of the same sentence.
In this subsection, we suppose every written character (instead of a word for English; \cref{sec:text-constituency-parsing-evaluation}) corresponds to a unit-length segment. 
We evaluate \structaiou by running a pre-trained Hebrew constituency parsing model \citep[\texttt{benepar\_he2};][]{kitaev-etal-2019-multilingual} on the SPMRL 2013 Hebrew development set \citep{seddah-etal-2013-overview}.
Similarly to the English text parsing evaluation result (\cref{sec:text-constituency-parsing-evaluation}), we obtain a high Spearman rank correlation coefficient of 0.823 (over 10-sentence buckets) between \structaiou (0.959 averaged across sentences) and \parseval F1 scores measured by EVALB-SPMRL (93.3).
    \begin{figure}[t]
	\centering
	\begin{minipage}{0.45\textwidth}
            \small
		\textbf{Ground-truth Tree} \\
  
            \centering
		\begin{forest} for tree={inner sep=0pt,l=10pt,l sep=6pt,s sep=11pt}
			[FRAGQ
					[\textcolor{cyan}{NP}
						[\textcolor{orange}{SYN\_WDT}				[\textcolor{blue}{DTT} [\<'yzh>]]]
						[\textcolor{magenta}{NP} [\textcolor{green}{SYN\_NN} [\textcolor{pink}{NN} [\<p.sw`ym>]]]]
					]
					[\textcolor{brown}{SYN\_yyQM} [\textcolor{olive}{yyQM} [?]]]
			]
		\end{forest}
	\end{minipage}
	\begin{minipage}{0.45\textwidth}
            \small 
		\textbf{Predicted Tree} \\
  
            \centering
		\begin{forest} for tree={inner sep=0pt,l=10pt,l sep=6pt,s sep=11pt}
			[SQ
					[\textcolor{cyan}{NP}
						[\textcolor{orange}{SYN\_WDT}				[
								\textcolor{blue}{DTT} [{\<'yzh>}]]
						]
						[\textcolor{magenta}{NP}
							[SYN\_NNT
								[BNT
										[\<p.sw`>]
								]
							]
							[NP
									[
										\textcolor{green}{SYN\_NN} [\textcolor{pink}{NN} [\<ym>] ]
									]
							]
						]
					]
					[\textcolor{brown}{SYN\_yyQM} [\textcolor{olive}{yyQM} [?]]]
			]
		\end{forest}
	\end{minipage}
	\caption{
		Parse trees of the Hebrew sentence ?\<p.sw`ym> \<'yzh> with two possible morphological analyses (\structaiou = 0.635). The plural morpheme \<yM> appears as a separate token in the predicted tree.
		Best viewed in color: the aligned nodes are shown in the same color.
	}
	\label{fig:hebrew-par-example}
\end{figure}

In addition, we demonstrate that \structaiou naturally provides a metric that supports misaligned morphological analysis.
The default tokenization in the SPMRL dataset does not extract the plural morphemes \<wt> and \<yM>; therefore, simply extracting the plural morphemes forms another acceptable tokenization strategy (see \cref{fig:hebrew-par-example} for an example).
We break the nouns ending with these two morphemes and feed the new tokenization to the \texttt{benepar\_he2} model.\footnote{The \texttt{benepar\_he2} model is not trained on this tokenization; however, we expect the model to work reasonably well, since it uses XLM-R \citep{conneau-etal-2020-unsupervised} as the word embeddings, which provides syntactic information of the new tokenization. \label{footnote:benepar-he2-xlm-r}}
The prediction with our new tokenization receives a \structaiou of 0.907 against the ground-truth---as desired, it is lower than 0.959 with the ground-truth tokenization.
However, the \structaiou score remains high, reflecting the facts that (1) the manipulation introduces misalignment between parses, and (2) the Benepar model is fairly robust to such mismatch on tokenization (see \cref{footnote:benepar-he2-xlm-r}).
In contrast to \textsc{TedEval} \citep{tsarfaty-etal-2012-joint}, which treats all mismatched nodes as errors with the same penalization in the final metric, \structaiou offers an alternative approach for evaluating parsing quality under misaligned morphological analyses, assigning partial credit to aligned same-label nodes with $\iou > 0$.

%% file: src/figs/02-speech-comparison.tex
\begin{figure}[t]
    \centering
    \includegraphics[width=0.48\textwidth]{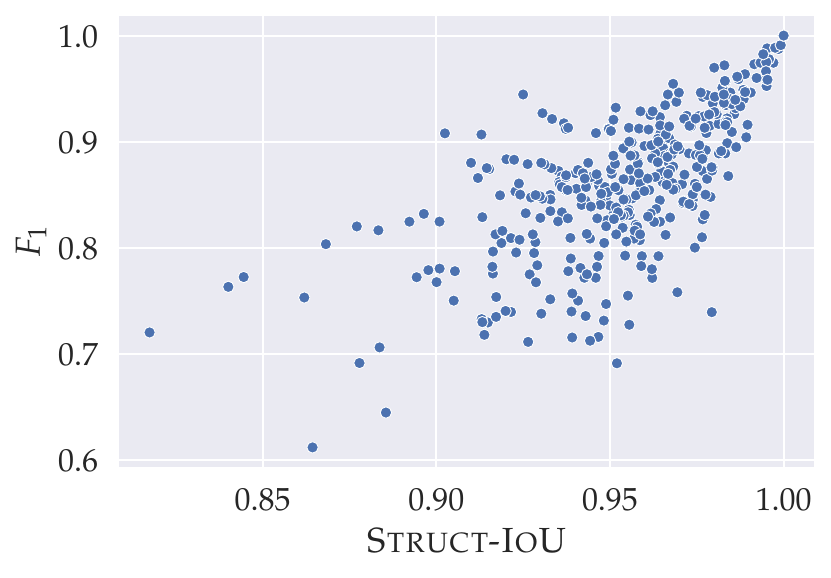}
    \caption{\label{fig:structa-iou-vs-parseval-speech}
    \structaiou vs. \parseval $F_1$ on NXT-SWBD (Spearman's correlation $\rho = 0.689$, p-value=$1.79\times 10^{-54}$). Each dot represents the results of the base model ($F_1$=85.4 on the full development set) on 10 random examples from the development set.}
\end{figure}

%% file: src/figs/03-speech-perturbation-example.tex
\begin{figure}[t]
    \centering
    \begin{subfigure}[t]{.45\textwidth}
        \centering
        \begin{tikzpicture}
            \draw[black,thick,] (0,0) -- (6,0)
            node[pos=0,mynode,fill=black,label=above:{$b_{i-1}^{(i-1)}$}]{}
            node[pos=0.4,mynode,fill=blue,label=above:{\textcolor{blue}{$b_i^{(i-1)}$}}]{}
            node[pos=1,mynode,fill=black,label=above:{$b_{i+1}^{(i-1)}$}]{}
            node[pos=0.56,mynode,fill=red,label=above:{\textcolor{red}{$b_{i}^{(i)}$}}]{}
            node[pos=0.7,mynode,fill=black]{}
            node[pos=0.2,mynode,fill=black]{}
            ;
            \coordinate (L) at (1.2, 0);
            \coordinate (R) at (4.2, 0);
            \draw[decorate,decoration={brace,amplitude=5pt,raise=2pt,mirror},yshift=0pt] (L) -- (R) node [midway, yshift=-15pt, xshift=0pt] {$[s_i, e_i]$};
        \end{tikzpicture}
        \caption{Noise-$\delta$, where $s_i = b_i^{(i-1)} - \delta \cdot \left(b_i^{(i-1)} - b_{i-1}^{(i-1)}\right)$ and $e_i = b_i^{(i-1)} + \delta \cdot \left(b_{i+1}^{(i-1)} - b_i^{(i-1)}\right)$ denote the most left and right possible position of $b_i^{(i)}$.}
        \label{subfig:noise-delta}
    \end{subfigure}
    \begin{subfigure}[t]{.45\textwidth}
        \centering \vspace{5pt}
        \begin{tikzpicture}
            \draw[black,thick,] (0,0) -- (6,0)
            node[pos=0,mynode,fill=black,label=above:{$b_{i-1}$}]{}
            node[pos=1,mynode,fill=black,label=above:{$b_i$}]{}
            node[pos=0.56,mynode,fill=red,label=above:{\textcolor{red}{$b_i'$}}]{};
        \end{tikzpicture}
        \caption{Insert-$\delta$, with $r_i < \delta$; otherwise $b_i'$ will not be inserted.}
        \label{subfig:add-delta}
    \end{subfigure}
    \begin{subfigure}[t]{.45\textwidth}
        \centering \vspace{5pt}
        \begin{tikzpicture}
            \draw[black,thick,] (0,0) -- (6,0)
            node[pos=0,mynode,fill=black,label=above:{$b_{i-1}$}]{}
            node[pos=1,mynode,fill=black,label=above:{$b_{i+1}$}]{}
            node[pos=0.4,mynode,fill=blue,label=above:{\textcolor{blue}{$b_i$}}]{};
        \end{tikzpicture}
        \caption{Delete-$\delta$, with $r_i < \delta$; otherwise $b_i$ will not be deleted.}
        \label{subfig:delete-delta}
    \end{subfigure}
    \caption{
        \label{fig:perturbation-example}
        Examples of three types of perturbation: when applicable, the added boundaries are shown in red and the deleted boundaries are shown in blue.
        Best viewed in color.
    }
\end{figure}

%% file: src/figs/04-speech-perturbation-results.tex
\begin{figure}[t]
    \centering
    \includegraphics[width=0.48\textwidth]{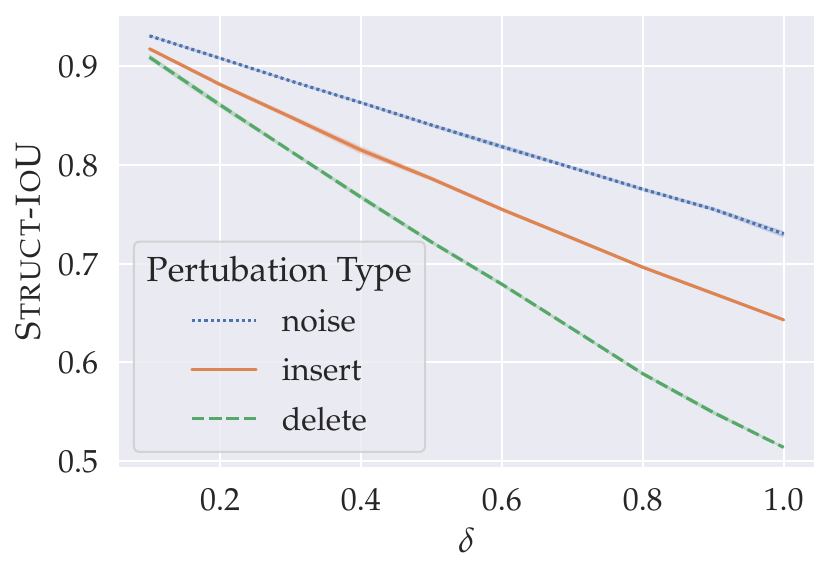}
    \caption{
        \label{fig:perturbation-speech} \structaiou scores with respect to $\delta$ for different types of perturbations.
    }
\end{figure}

%% file: src/figs/05-speech-structaiou-with-length.tex
\begin{figure}[t]
    \centering
    \includegraphics[width=0.48\textwidth]{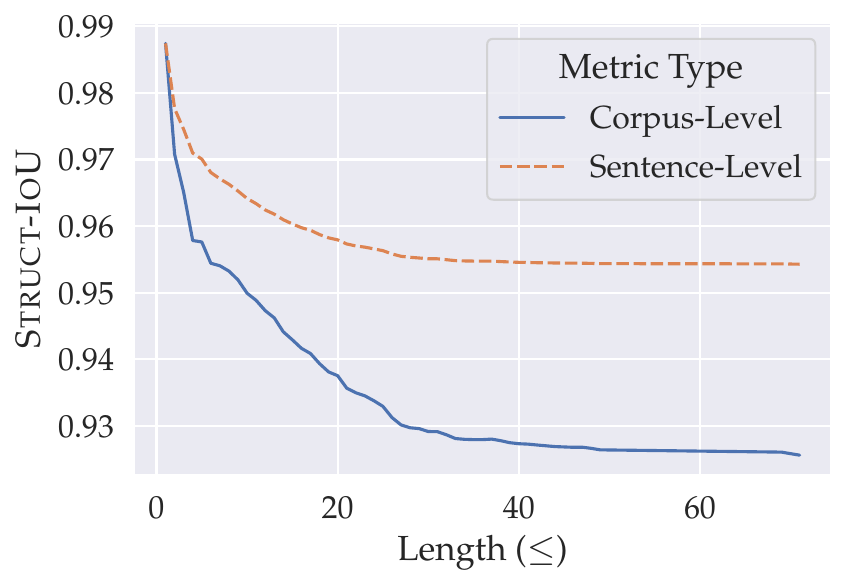}
    \caption{
        \label{fig:structa-iou-with-length} Corpus-level and sentence-level \structaiou scores of the predicted parse trees of the base model ($F_1 = 85.4$ on the development set), evaluated on development examples with less than or equal to a certain number of spoken words.
    }
\end{figure}

%% file: src/figs/06-text-comparison.tex
\begin{figure}[t]
    \centering
    \includegraphics[width=0.48\textwidth]{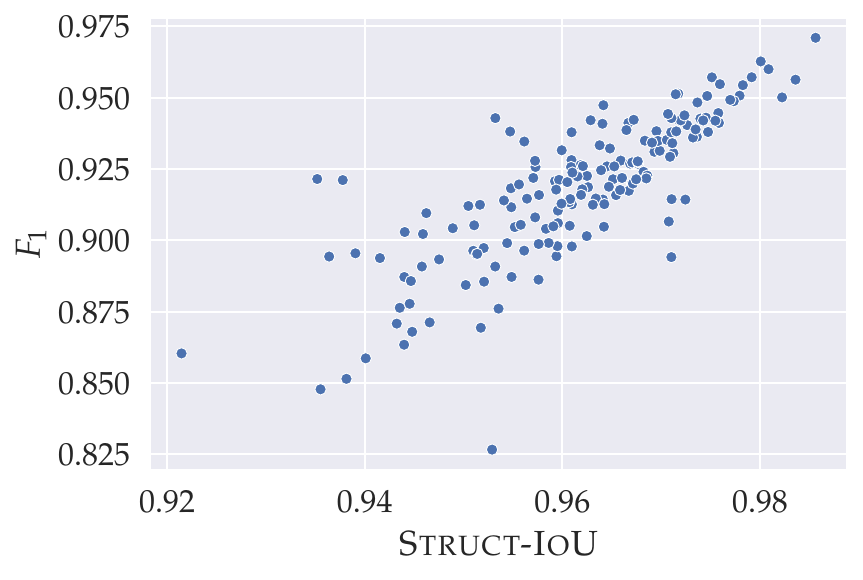}
    \caption{\label{fig:structa-iou-vs-parseval-text}
    Comparison of \structaiou and \parseval $F_1$ (Spearman's rank correlation $\rho = 0.821$, p-value=$8.16\times 10^{-43}$). Each dot represents the results of the base model on 10 random examples from the PTB development set.}
\end{figure}

%% file: src/figs/07-syntactic-ambiguity.tex
\begin{figure}[t]
    \centering \small
    \begin{forest} for tree={inner sep=0pt,l=10pt,l sep=10pt,s sep=11pt}
        [S
            [NP
                [DT [The]]
                [NN [girl]]
            ]
            [VP
                [VBD [saw]]
                [NP
                    [DT [a]]
                    [NN [cat]]
                    [PP
                        [P [with]]
                        [NP
                            [DT [a]]
                            [NN [telescope]]
                        ]
                    ]
                ]
            ]
        ]
    \end{forest}

    \vspace{-30pt}

    \begin{forest} for tree={inner sep=0pt,l=10pt,l sep=10pt,s sep=11pt}
        [S
            [NP
                [DT [The]]
                [NN [girl]]
            ]
            [VP
                [VBD [saw]]
                [NP
                    [DT [a]]
                    [NN [cat]]
                ]
                [PP
                    [P [with]]
                    [NP
                        [DT [a]]
                        [NN [telescope]]
                    ]
                ]
            ]
        ]
    \end{forest}
    \caption{\label{fig:example-ambiguous-sentences} An example syntactically ambiguous sentence: \textit{The girl saw a cat with a telescope}.
    Both parses are syntactically valid, but the first one implies that \textit{a cat} was holding the telescope, whereas the second implies \textit{the girl} was using the telescope.}
\end{figure}

\begin{figure}[t]
    \centering \small
    \begin{forest} for tree={inner sep=0pt,l=10pt,l sep=6pt,s sep=11pt}
        [NP
            [NP [N]]
            [PP
                [P]
                [NP
                    [NP [N]]
                    [PP
                        [P]
                        [NP [N]]
                    ]
                ]
            ]
        ]
    \end{forest}
    \hspace{10pt}
    \begin{forest} for tree={inner sep=0pt,l=10pt,l sep=6pt,s sep=11pt}
        [NP
            [NP
                [NP [N]]
                [PP
                    [P]
                    [NP
                        [N]
                    ]
                ]
            ]
            [PP
                [P]
                [NP [N]]
            ]
        ]
    \end{forest}
    \caption{\label{fig:example-ambiguous-sentences-synthetic} Two syntactically plausible parses of the \texttt{N (P N)\{2\}}, where NP denotes a noun phrase, and PP denotes a prepositional phrase.}
\end{figure}

%% file: src/figs/08-syntactic-ambiguity-results.tex
\begin{table}[t]
    \centering
    \small
    \begin{tabular}{lc}
        \toprule
        \bf Metric & \bf Ground-Truth vs. Random, Average \\
        \midrule
        \parseval $F_1$ & 27.3\\
        \structaiou & 61.9 \\
        \midrule
        & \bf Ground-Truth vs. Plausible, Lowest \\
        \midrule
        \parseval $F_1$ & 12.5\\
        \structaiou & 63.6 \\
        \bottomrule
    \end{tabular}
    \caption{
        \label{tab:syntactic-ambiguity-results}
        Average \parseval $F_1$ and \structaiou scores between the ground truth and a random binary tree, and the lowest possible scores between the ground truth and another syntactically plausible tree.
        Experiments are done on the string ``\texttt{N \text{(}P N\text{)}}\{\texttt{8}\}''.
        For simplicity, we report the unlabeled scores, where all nonterminals are treated as having the same label.
    }
\end{table}

%% file: src/60-discussion.tex
\section{Conclusion and Discussion}
In this paper, we present \structaiou, the first metric that computes the similarity between two parse trees over continuous spoken word boundaries.
\structaiou enables the evaluation of textless speech parsing \citep{lai-etal-2023-audio,tseng-etal-2023-cascading}, where no text or speech recognizer is used or available to parse spoken utterances.

In the canonical text and speech parsing settings, \structaiou complements the existing evaluation metrics \citep{black-etal-1991-procedure,roark-etal-2006-sparseval,tsarfaty-etal-2012-joint}.
Even for the evaluation of English constituency parsing, \structaiou shows a higher tolerance to potential syntactic ambiguity under certain scenarios, providing an alternative interpretation of the parsing quality.

Faithful evaluation of parsing quality is crucial for developing both speech and text parsing models.
In supervised parsing, it has been common sense that higher evaluation metric scores (i.e.,  \parseval $F_1$) imply better models.
However, the misalignment between linguistically annotated ground truths and model predictions, especially unsupervised parsing model predictions, does not necessarily indicate poor parsing quality of the models \citep{shi-etal-2020-role}---instead, the models may have learned different but equally valid structures.
Conversely, annotations made by linguistic experts (such as the Penn Treebank) may exhibit discrepancies when compared to the responses of native speakers who lack formal linguistic training.
We suggest that future work investigate what properties of the parses are emphasized by each evaluation metric, and consider multi-dimensional evaluation metrics \interalia{kasai-etal-2022-bidimensional}.

\section*{Acknowledgement}
We thank Yudong Li, Jiayuan Mao, and the anonymous reviewers for their valuable suggestions. 
This work is supported in part by a Google Ph.D. Fellowship to FS. 

\section*{Limitations}
\structaiou is designed to evaluate constituency parse trees over continuous spoken word boundaries, and is not directly applicable to evaluate other types of parses, such as dependency parse trees; however, it may be extended to evaluate other types of parse trees by modifying the alignment constraints.
We leave the extension of \structaiou to other types of parses as future work.
We do not foresee any risk associated with the use of \structaiou beyond the inherent minimal risks encountered in computer science research.

%% file: src/99-appendix.tex
\noindent \textbf{\Large Appendix}
\section{Proof of Corollaries and Propositions}
\label{appendix:proof}
We present the proof of the corollaries mentioned in the main content as follows.

\noindent \textbf{\cref{corollary:segment-tree-parent-cover}}
    For nodes $\bp, \bn$, if $\bp \in C_\bn$, then $I_\bp \subseteq I_\bn$.
\begin{proof}
    According to the definition of open intervals and \cref{def:segment-tree-node} (3),
    \begin{align*}
        & a_\bn \leq a_\bp < b_\bp \leq b_\bn \\
        \Rightarrow & I_\bp = (a_\bp, b_\bp) \subseteq (a_\bn, b_\bn) = I_{\bn}.
    \end{align*}
\end{proof}

\noindent \textbf{\cref{corollary:ancestor-cover}}
If node $\bp$ is an ancestor of node $\bq$, then $I_\bp \supseteq I_\bq$.
\begin{proof}
    According to Definition~\ref{def:ancestor}, there exists a sequence of nodes $\bn_0, \bn_1, \ldots, \bn_k (k\geq 1)$ such that (1) $\bn_0 = \bp$, (2) $\bn_k = \bq$ and (3) for any $i \in [k]$, $\bn_{i} \in \mathbf{C}_{\bn_{i-1}}$.

    Corollary~\ref{corollary:segment-tree-parent-cover} implies that for any $i \in [k]$, $I_{\bn_{i-1}} \supseteq I_{\bn_i} \Rightarrow I_{\bn_0}\supseteq I_{\bn_k}\Rightarrow I_\bp \supseteq I_\bq$.
\end{proof}

\noindent\textbf{\cref{corollary:segment-tree-characterization}}
    A relaxed segment tree can be uniquely characterized by its root node.
\begin{proof}
    $(\Rightarrow)$ Definition~\ref{def:ex-segment-tree} implies that each relaxed segment tree has one root node.

    $(\Leftarrow)$ Given a specific node $\bn$, we have the unique set $\mathcal{N} = \{\bn\} \cup \{\bn': \bn' \text{ is a descendant of } \bn\}$, and therefore extract the set of all nodes in the relaxed segment tree rooted at $\bn$.
\end{proof}

\noindent\textbf{\cref{proposition:non-ancestor-disjoint}}
    For a relaxed segment tree $\cT$ and $\bp, \bq \in N_\cT$,  $\bp$ is neither an ancestor nor a descendant of $\bq$ $\Leftrightarrow$ $I_\bp \cap I_\bq = \emptyset$.
\begin{proof}
    ($\Rightarrow$) Let $\bz$ denote the least common ancestor of $\bp$ and $\bq$. There exists $\bp', \bq' \in C_\bz (\bp' \neq \bq')$ such that $I_{\bp'} \supseteq I_\bp$ and $I_{\bq'} \supseteq I_\bq$; therefore
    \begin{align*}
    I_p \cap I_q \subseteq I_{\bp'} \cap I_{\bq'}
    \overset{\text{Definition~\ref{def:segment-tree-node} (2)}}{=}\emptyset
    \Rightarrow I_\bp \cap I_\bq = \emptyset.
    \end{align*}
    \\
    ($\Leftarrow$) If $I_{\bp} \cap I_{\bq} = \emptyset$, according to \cref{def:segment-tree-node} (3) and \cref{def:ancestor}, $\bp$ is not an ancestor of $\bq$ and vice versa.
\end{proof}

\noindent\textbf{\cref{cor:ordered-disjoint-descendant-sequence}}
    Given an ordered disjoint descendant sequence $\mathbf{S} = \left(\bn_1, \bn_2, \ldots, \bn_k\right)$ of $\bn$, for any $i \in [k-1]$, $b_{\bn_i} \leq a_{\bn_{i+1}}$.
\begin{proof}
    If there exists $i \in [k-1]$ such that $b_{\bn_i} > a_{\bn_{i+1}}$, then
    \begin{align*}
        & I_{\bn_i}\cap I_{\bn_{i+1}}  \\
        = & (a_{\bn_i}, b_{\bn_i}) \cap (a_{\bn_{i+1}}, b_{\bn_{i+1}}) \\
        = & \left\{x: \max(a_{\bn_i}, a_{\bn_{i+1}}) < x < \min(b_{\bn_i}, b_{\bn_{i+1}})\right\} \\
        = & \left\{x: a_{\bn_{i+1}} < x < \min(b_{\bn_i}, b_{\bn_{i+1}})\right\} \\
        & \quad \textit{(Definition~\ref{def:ordered-disjoint-descendant-sequence} (1)).}
    \end{align*}
    Since $b_{\bn_{i+1}} > a_{\bn_{i+1}}$ (definition of open intervals), \begin{align*}
        & a_{\bn_{i+1}} < \min(b_{\bn_i}, b_{\bn_{i+1}})
        \Rightarrow I_{\bn_i}\cap I_{\bn_{i+1}} \neq \emptyset.
    \end{align*}
    This conflicts with Definition~\ref{def:ordered-disjoint-descendant-sequence} (2).
\end{proof}